\documentclass[runningheads]{llncs}

\usepackage{eccv}

\usepackage{eccvabbrv}

\usepackage{graphicx}
\usepackage{booktabs}
\usepackage{multirow}
\usepackage{wrapfig}
\usepackage{diagbox}
\usepackage{makecell}
\usepackage{pifont}
\usepackage{xcolor}
\usepackage{colortbl}

\usepackage[accsupp]{axessibility}  

\usepackage{hyperref}
\hypersetup{
urlcolor=red,
}

\newlength\savewidth

\newcommand{\cmark}{\ding{51}}

\def\ie{\textit{i.e.}}

\usepackage[capitalize]{cleveref}
\crefname{section}{Sec.}{Secs.}
\Crefname{section}{Section}{Sections}
\Crefname{table}{Table}{Tables}
\crefname{table}{Tab.}{Tabs.}

\usepackage{orcidlink}

\begin{document}

\title{Reason2Drive: Towards Interpretable and Chain-based Reasoning for Autonomous Driving}

\titlerunning{Reason2Drive}

\author{Ming Nie\inst{1} \and
Renyuan Peng\inst{1} \and
Chunwei Wang\inst{2} \and
Xinyue Cai\inst{2} \and
Jianhua Han\inst{2} \and
Hang Xu\inst{2} \and
Li Zhang\inst{1}\thanks{Li Zhang (lizhangfd@fudan.edu.cn) and Hang Xu (chromexbjxh@gmail.com) are the corresponding authors.}
}

\authorrunning{Ming Nie, et al.}

\institute{School of Data Science, Fudan University \and
Huawei Noah’s Ark Lab \\
\vspace{4mm}
\href{https://github.com/fudan-zvg/reason2drive}{\textcolor{magenta}{https://github.com/fudan-zvg/reason2drive}}
}

\maketitle

\begin{abstract}
Large vision-language models (VLMs) have garnered increasing interest in autonomous driving areas, due to their advanced capabilities in complex reasoning tasks essential for highly autonomous vehicle behavior.
Despite their potential, research in autonomous systems is hindered by the lack of datasets with annotated reasoning chains that explain the decision-making processes in driving.
To bridge this gap, we present Reason2Drive, a benchmark dataset with over 600K video-text pairs, aimed at facilitating the study of interpretable reasoning in complex driving environments.
We distinctly characterize the autonomous driving process as a sequential combination of \textit{perception}, \textit{prediction}, and \textit{reasoning} steps, and the question-answer pairs are automatically collected from a diverse range of open-source outdoor driving datasets, including nuScenes, Waymo and ONCE.
Moreover, we introduce a novel aggregated evaluation metric to assess chain-based reasoning performance in autonomous systems, addressing the reasoning ambiguities of existing metrics such as BLEU and CIDEr.
Based on the proposed benchmark, we conduct experiments to assess various existing VLMs, revealing insights into their reasoning capabilities.
Additionally, we develop an efficient approach to empower VLMs to leverage object-level perceptual elements in both feature extraction and prediction, further enhancing their reasoning accuracy.
Extendable experiments demonstrate the supportive effect of Reason2Drive towards visual reasoning and downstream planning tasks.
\end{abstract}
\vspace{-2mm}

\section{Introduction}
\begin{wrapfigure}{r}{6cm}
    \centering
    \includegraphics[scale=0.28]{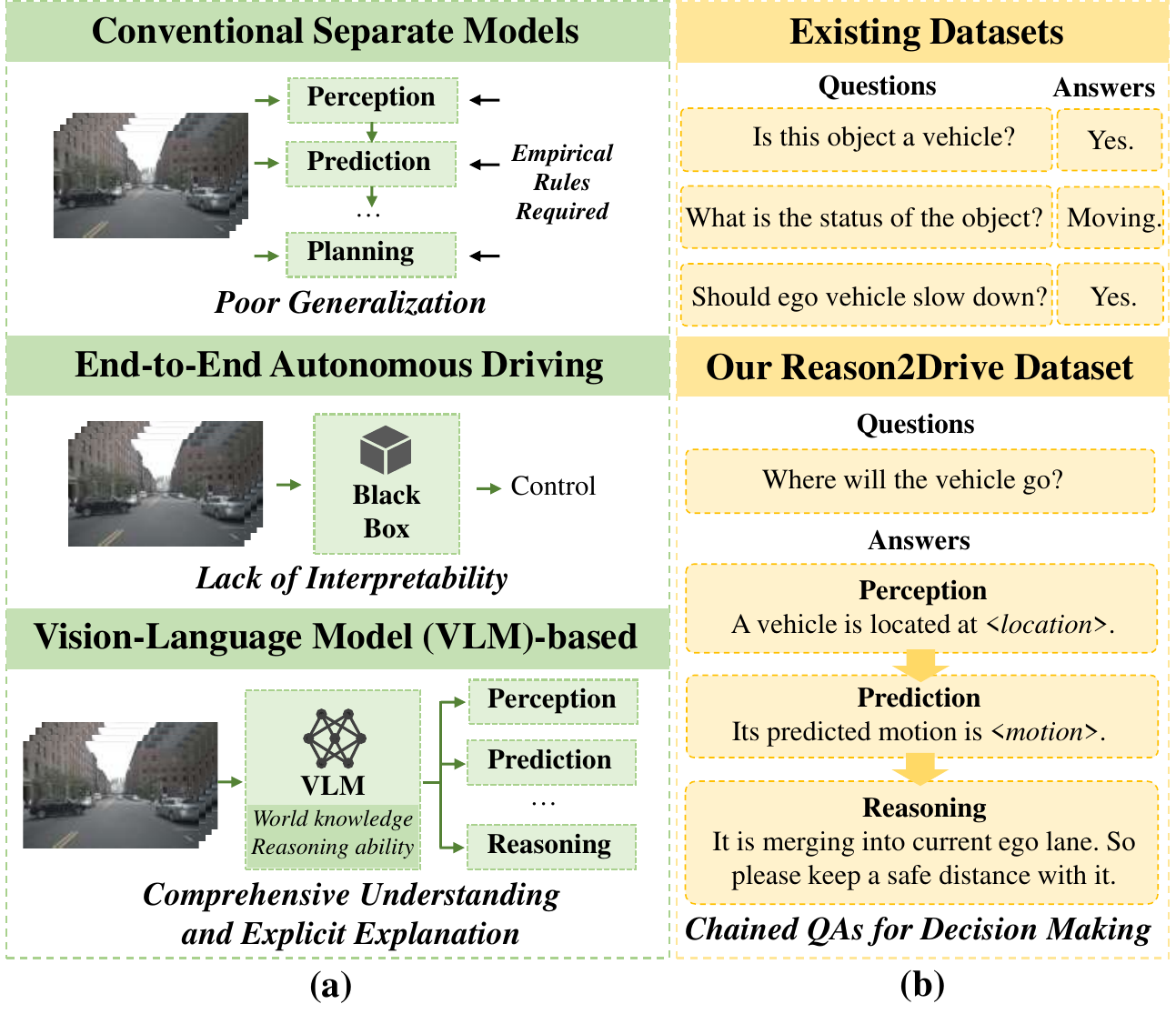}
    \caption{(a) Different decision-making processes in autonomous driving. 
    (b) Language-based dataset comparison.}
    \label{fig-intro}
    \vspace{-20pt}
\end{wrapfigure}

Modern autonomous driving systems face challenges related to generalization issues across diverse scenarios, which is often attributed to the reliance on empirical and intricate rules involved in decision-making. 
To reduce dependence on such rules, recent end-to-end approaches~\cite{hu2023planning, chen2023end} have been developed to derive control signals directly from sensor inputs, treating the system as a black box that requires extensive data for training.
However, this approach tends to obscure the underlying logic of decisions, complicating failure diagnosis in real-world applications. 
In contrast, Large Vision-Language Models (VLMs) offer a promising alternative, potentially enhancing interpretability and generalization for these systems.
With their broad world knowledge and advanced reasoning abilities, as illustrated in Fig.~\ref{fig-intro}(a), VLMs have the potential to provide a more thorough understanding and explicit explanation for reliable decision-making.
Nonetheless, existing works~\cite{qian2023nuscenes,wu2023referring} primarily focused on the straightforward adaptation of question-answering tasks to the autonomous driving; how to exploit VLMs to facilitate the reasoning abilities of autonomous systems is still under exploration.

One reason that hinders the research in this field lies in the scarcity of datasets, especially those chained-based reasoning labels that elucidate the decision-making process.
Most existing datasets~\cite{deruyttere2022talk2car,qian2023nuscenes,wu2023language} often oversimplify the complex processes of driving into straightforward question-answering tasks with only a few specific tasks covered.
As depicted in Fig.~\ref{fig-intro}(b), they typically provide closed-form annotations constrained to boolean (i.e., yes or no) answers or limited multiple-choice responses (e.g., stopped, parked, and moving).
However, autonomous driving transcends a simplistic QA process.
It encompasses a multi-step approach involving \textit{perception}, \textit{prediction}, and \textit{reasoning}, each of which plays an indispensable role in the decision-making.
Therefore, it is crucial to introduce a novel benchmark annotated with detailed decision-making reasoning for assessing the reasoning abilities of current VLMs.

To this end, we introduce Reason2Drive, a new benchmark comprising over 600K video-text pairs, characterized by intricate driving instructions and a series of perception, prediction and reasoning steps. 
Our benchmark builds upon widely-used open-source driving datasets including nuScenes~\cite{caesar2020nuscenes}, Waymo~\cite{sun2020scalability}, and ONCE~\cite{2021Once}, utilizing an extensible annotation schema. 
Specifically, we structure the comprehensive annotations into object-centric database and integrate it into manually crafted templates to create paired data for VLMs at both object and scenario levels.
To enhance diversity, GPT-4 and manual instructions are employed for verification and enrichment purposes.
Notably, Reason2Drive is the most extensive dataset available to date, outperforming existing datasets in scale and the complexity of reasoning chains included, which is a distinctive attribute not present in other datasets.

Furthermore, we observe a fundamental flaw in the current evaluation of VLMs on autonomous driving tasks, due to the inherent reasoning ambiguities of traditional caption-based metrics like BLEU~\cite{papineni2002bleu} and CIDEr~\cite{vedantam2015cider}.
These metrics mainly measure text generation from a holistic perspective, without considering the causal relationship between the reasoning steps and the final conclusion.
For example, when the VLM suggests ego vehicle turning left, we cannot ascertain from these metrics whether its reasoning steps effectively support the final decision.
To address this issue, we propose a new aggregated evaluation metric, ADRScore, specifically designed to measure chain-based reasoning performance in autonomous systems, which aims to resolve the reasoning ambiguities associated with current metrics.

Utilizing the proposed benchmark, we undertake experiments to assess various existing VLMs, thereby unveiling valuable insights into their reasoning capabilities.
We find that most methods struggle to effectively leverage perceptual priors, resulting in subpar reasoning performance.
Additionally, constrained by the language model functioning solely as a decoder, these methods often fail to deliver accurate perceptual results, which is a crucial component for verifying a model's spatial reasoning capability.
To alleviate this dilemma, we present a simple yet efficient framework, 
augmenting existing VLMs with two new components: a prior tokenizer and an instructed vision decoder, which aim to bolster the models' visual localization capabilities within the encoder and decoder, respectively.
Extendable experiments demonstrate the supportive effect of Reason2Drive towards visual reasoning and downstream planning tasks.

The contributions of this paper are summarized as follows:
\textbf{(i)}
We publish a novel visual instruction tuning dataset aimed at facilitating interpretable and chain-based reasoning autonomous systems.
\textbf{(ii)}
We introduce a novel evaluation metric, ADRScore, to assess chain-based reasoning performance in autonomous driving, effectively addressing the reasoning ambiguities present in existing metrics.
\textbf{(iii)}
We conduct experiments to assess a range of existing VLMs, revealing valuable insights into their reasoning capabilities.
\textbf{(iv)}
To address the challenges posed by inefficient prior feature extraction and inaccurate perceptual predictions, we introduce an efficient approach to integrate these elements into VLMs.
This results in a substantial improvement in reasoning accuracy and provides remarkable support for downstream planning tasks.
Our method surpasses all baselines, notably achieving impressive generalization in unseen scenarios.

\section{Related Work}
\noindent{\textbf{Multimodal large language model.}}
The current state of large language models provides remarkable abilities in natural language understanding and generation (\cite{chowdhery2022palm,touvron2023llama,chiang2023vicuna,2023gpt4}).
Inspired by the potential of large language models, a multitude of multimodal models has emerged, intended to enhance these models' capabilities in achieving multi-modal comprehension.
Blip-2~\cite{li2023blip} aligns visual and language features by utilizing a learnable Q-former. 
LLaVA~\cite{liu2023visual} and MiniGPT-4~\cite{zhu2023minigpt} initially align image-text features and then proceed with instruction tuning. 
Additionally, Video-LLaMA~\cite{zhang2023video} and ImageBind-LLM~\cite{han2023imagebind} integrate multiple modalities into the input, aligning features from various sources like images, videos, audio, and point clouds, consolidating them into the space of language features.
Kosmos-2~\cite{peng2023kosmos} and Shikra~\cite{chen2023shikra} perform object detection based on instructions and also accomplish grounded visual question answering.
DetGPT~\cite{pi2023detgpt} connects a fixed multi-modal LLM with a customizable detector based on user instructions. 
LISA~\cite{lai2023lisa} efficiently embeds segmentation abilities into multi-modal LLMs, showcasing self-reasoning for current perception systems.
The previous works have demonstrated that current large-scale multimodal models can achieve cross-modal alignment, enabling comprehension and inference towards images and more.
These models can not only perform perceptual tasks like detection but also accomplish preliminary reasoning tasks.

\begin{figure*}[t]
    \begin{center}
        \includegraphics[width=1.0\linewidth]{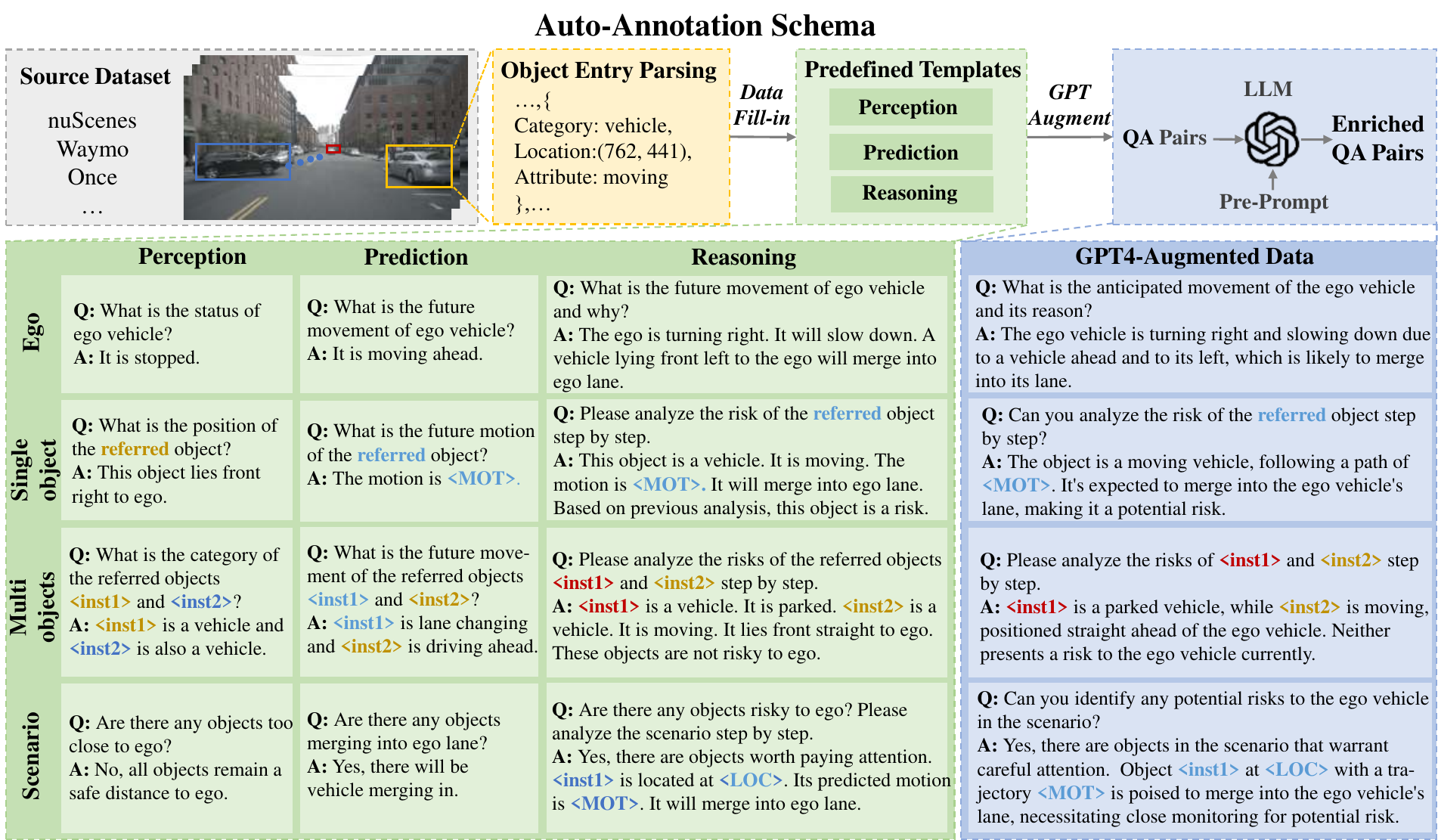}
    \end{center}
    \vspace{-4mm}
    \caption{Schema of Our Reason2Drive Dataset. The upper part illustrates the pipeline for the automated construction of datasets. 
    The lower part shows detailed instances of perception, prediction, and reasoning, accompanied by outcomes after applying GPT-4 for data augmentation. 
    The special tokens hold distinct definitions: \texttt{<Inst*>} represents a specified instance, \texttt{<MOT>} signifies a forecasted sequence of trajectory coordinates, and \texttt{<LOC>} denotes positional coordinates.
    The colors associated with these tokens correspond to the highlighted objects in the upper-left image's boxes.}
    \label{fig-schema}
    \vspace{-7mm}
\end{figure*}

\noindent{\textbf{Vision language tasks in autonomous driving.}}
Currently, VLMs have demonstrated robust capabilities in scene perception and understanding.
Extensive efforts have been dedicated to the realm of autonomous driving, leveraging VLM to achieve comprehensive scene understanding and perform diverse tasks~\cite{mao2023gpt, fu2023drive, xu2023drivegpt4, ding2023hilm}.
Simultaneously, substantial works are in progress to create datasets and models tailored to various tasks.
Talk2Car~\cite{deruyttere2022talk2car} proposes the first object referral dataset for grounding commands for self-driving cars in free natural language into the visual context.
But it exclusively contains information about visible objects.
While DRAMA~\cite{malla2023drama} outlines the overall scene risk, it lacks precise perception annotation.
NuPrompt~\cite{wu2023language} and Refer-KITTI~\cite{wu2023referring} offer language prompt sets for driving scenes but primarily concentrate on multi-object tracking tasks.
NuScenesQA~\cite{qian2023nuscenes} and DriveLM~\cite{drivelm2023} build visual question-answering (VQA) datasets for scenario understanding. 
However, their primary emphasis is on the perceptual information in the scene, lacking annotations for the analysis and complex reasoning of the entire scenario.
To address the limitations of existing works, we construct a thorough dataset covering perception, prediction, and complex reasoning, 
additionally with an improved vision-language model for better analyzing autonomous driving scenarios.

\section{Reason2Drive Dataset}
We introduce Reason2Drive, a dataset that comprises comprehensive driving instructions and a chain-based reasoning framework for decision-making. Our dataset is characterized by the following key aspects:
\vspace{2mm}
\begin{itemize}
    \item \textbf{Quantity}: It stands out as the largest language-based driving dataset available, collated from prominent publicly accessible datasets worldwide.
    \item \textbf{Quality}: Reason2Drive offers a more precise representation of driving activities, including \textit{perception}, \textit{prediction} and \textit{reasoning}, with a reliable auto-annotation schema for data collection.
    \vspace{2mm}
    \item \textbf{Diversity}: 
    (i) The dataset exhibits a broader range of scenes, encompassing both object-level and scenario-level data. This diversity includes object types, visual and motion attributes, object locations, and relationships relative to the ego-vehicle. (ii) It includes more intricate question-answer pairs, enhanced by GPT-4, along with longer text passages featuring step-by-step reasoning.
    \vspace{2mm}
    \item \textbf{Protocols}: A novel evaluation metric is introduced to assess the reasoning capabilities of VLMs. Different from those widely used in the NLP community, it takes into account not only perception results but also reasoning ambiguities, providing a more holistic evaluation of the VLM's reasoning capacity for autonomous driving scenarios.
\end{itemize}

Further details regarding the data collection process, statistical data analysis, and benchmark protocols are provided in the subsequent section.

\makeatletter 
    \newcommand\figcaption{\def\@captype{figure}\caption} 
    \newcommand\tabcaption{\def\@captype{table}\caption} 
\makeatother
\renewcommand\arraystretch{1.3}

\definecolor{Talk2Car}{RGB}{237, 125, 49}
\definecolor{CityFlow-NL}{RGB}{255, 192, 203}
\definecolor{CARLA-NAV}{RGB}{255, 192, 0}
\definecolor{NuScenes-QA}{RGB}{91, 155, 213}
\definecolor{Refer-KITTI}{RGB}{112, 173, 71}
\definecolor{Reason2Drive}{RGB}{255, 0, 0}
\definecolor{DRAMA}{RGB}{112, 48, 160}
\definecolor{Rank2Tell}{RGB}{102, 192, 249}

\begin{figure*}[t]
\begin{minipage}[b]{0.55\textwidth}
    \centering
    \resizebox{0.9\linewidth}{!}{
    \begin{tabular}{c|cc}
    \midrule[1.0pt]
    \rowcolor[gray]{0.8}Dataset & Description & Source Datasets \\
    \midrule[1.0pt]
    Talk2Car~\cite{deruyttere2022talk2car} \textcolor{Talk2Car}{$\blacksquare$} & Object referral & nuScenes \\
    CityFlow-NL~\cite{feng2021cityflow} \textcolor{CityFlow-NL}{$\blacksquare$} & Tracking \& retrieval & CityFlow \\
    CARLA-NAV~\cite{jain2023ground} \textcolor{CARLA-NAV}{$\blacksquare$} & Segmentation \& prediction & CARLA Simulator \\
    NuPrompt~\cite{wu2023language} $\blacksquare$ & Multi-object tracking & nuScenes \\
    NuScenes-QA~\cite{qian2023nuscenes} \textcolor{NuScenes-QA}{$\blacksquare$} & Perception & nuScenes \\
    Refer-KITTI~\cite{wu2023referring} \textcolor{Refer-KITTI}{$\blacksquare$} & Multi-object tracking & KITTI \\
    Talk2BEV~\cite{dewangan2023talk2bev} $\blacksquare$ & Visual understanding & nuScenes \\
    DRAMA~\cite{malla2023drama} \textcolor{DRAMA}{$\blacksquare$} & Risk localization & self-collected \\
    Rank2Tell~\cite{sachdeva2023rank2tell} \textcolor{Rank2Tell}{$\blacksquare$} & Risk localization \& ranking & self-collected \\
    \midrule[1.0pt]
    \multirow{3}{*}{Reason2Drive\textcolor{Reason2Drive}{$\blacksquare$}}  & Perception & nuScenes \\
    \specialrule{0em}{-1pt}{-1pt} & Prediction & Waymo \\
    \specialrule{0em}{-1pt}{-1pt} & Reasoning & ONCE \\
    \bottomrule
    \end{tabular}}
    \tabcaption{The comparison between our Reason2Drive dataset and other prompt-based datasets. $\blacksquare$ means dataset not published.}
    \label{tab-bench}
    \vspace{1mm}
\end{minipage}
\hspace{3mm}
\begin{minipage}[b]{0.4\textwidth}
    \centering
    \includegraphics[scale=0.2]{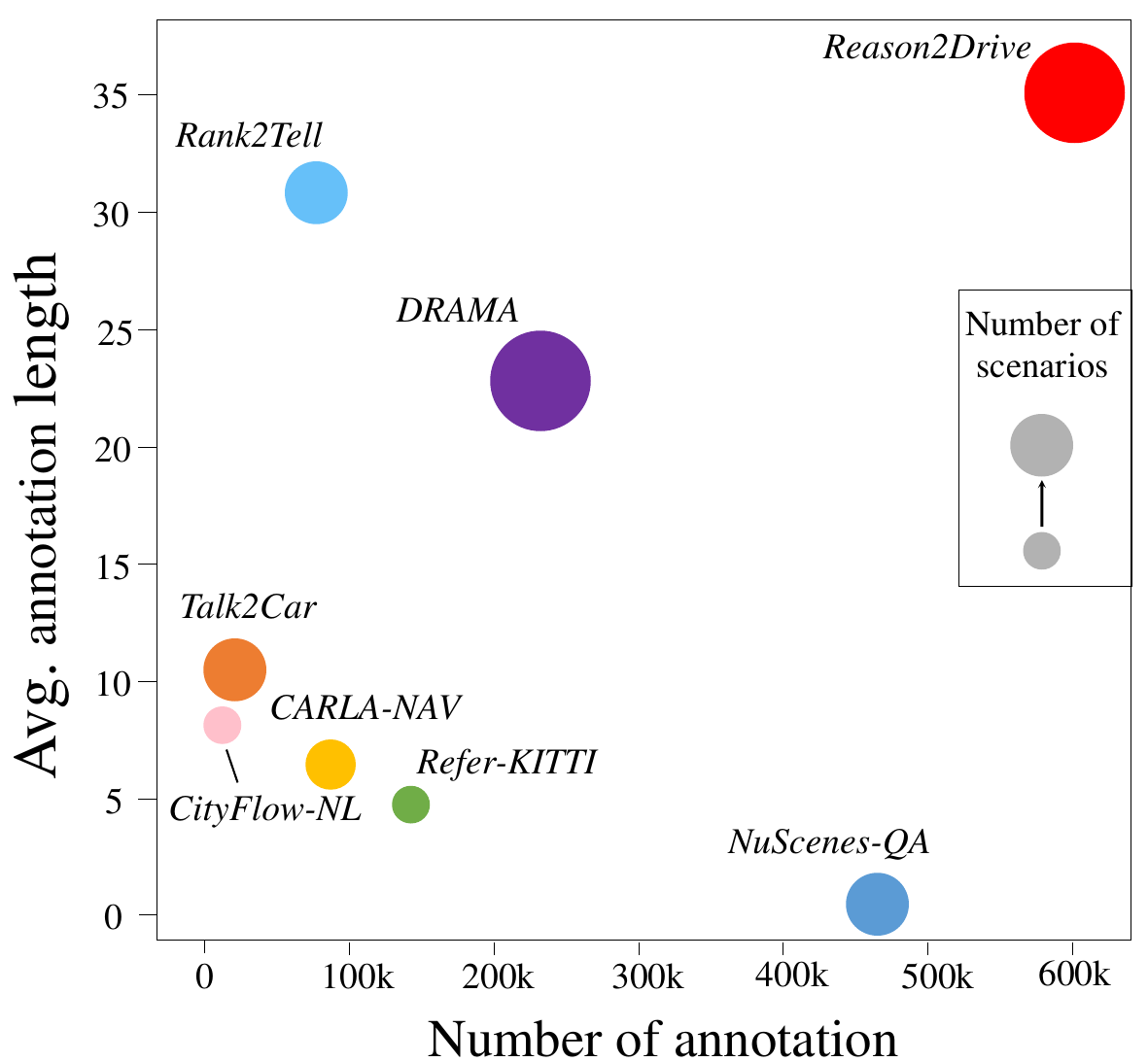}
    \figcaption{Data quality comparison. Reason2Drive is larger in scale, richer in data content, and more diverse in scenarios.}
    \label{fig-scatter}
\end{minipage}
\vspace{-5mm}
\end{figure*}

\subsection{Dataset collection} \label{dataset collection}

As illustrated in Fig.~\ref{fig-schema}, we employ an extensible annotation schema, constructing data in the forms of question-answer pairs.
Specifically, we first leverage a diverse array of publicly available datasets collected in different regions worldwide, including nuScenes, Waymo, and ONCE, and then parse their comprehensive annotations into object-centric database.
The database is organized in key frames, and each frame stores object entries containing various details pertaining to its driving actions, including location, category, attributes and more.
Afterwards, the structured database is required to answer the manually crafted question templates, which are divided into different tasks (i.e., perception, prediction and reasoning) at both object-level and scenario-level.
Subsequently, GPT-4 is involved for verification and enrichment purposes.
The example of GPT augmented data and manual instructions are provided in the appendix.

Due to the complexity of autonomous driving activities, we categorize the tasks into three distinct groups to acquire diversified data: perception, prediction and reasoning.
The specifics and distinctions of these three types of tasks are elaborated as follows:

\begin{itemize}
    \item \textbf{Perception task} is designed to identify objects within the driving scenario, assessing the fundamental perceptual capabilities of VLMs in outdoor environments.
    \item \textbf{Prediction task} entails the prediction of future states of key objects within the perceptual range, challenging VLMs to infer the intentions of objects with video input.
    \item \textbf{Reasoning task} prompts the analysis of the current perceptual and predicted states step by step, requiring the deduction of reasoned inferences and decisions through a chain of thoughts (COT) approach.
\end{itemize}

For each task, we further categorize the data into object-level and scenario-level.
In more detail, 

\begin{itemize}
    \item \textbf{Object-level} data is formatted to benchmark the foundational capabilities of specific objects. As for perception, we address the location and attributes of objects such as moving status and distance to ego, while for prediction, future motion and merging-in/out status are considered.
    \item \textbf{Scenario-level} data is organized from a global perspective towards driving environment and ego-driving instructions.
    It focuses on whether there is an object worth noting currently (perception), whether there is an object worth noting in the future (prediction) and why (reasoning).
    For example, as illustrated in Fig.~\ref{fig-schema}, models are asked to identify distances, merging states and other risks from the whole scene.
    It verifies the agent's ability to perceive the entire driving scene rather than specifying objects, thus more challenging.
\end{itemize}


\begin{figure*}[t]
\begin{minipage}[b]{0.4\textwidth}
    \centering
    \includegraphics[scale=0.18]{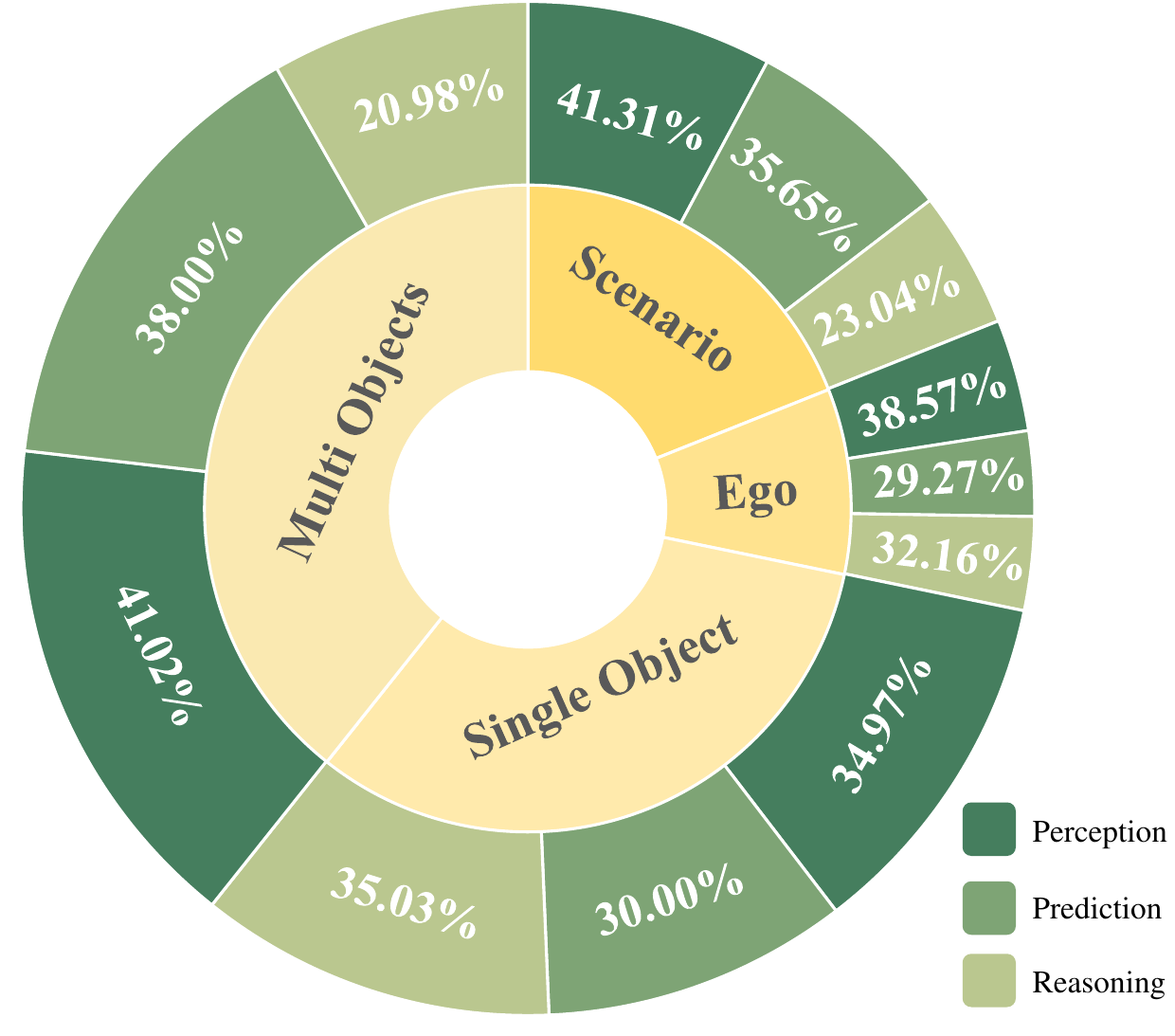}
    \figcaption{Statistical distribution of different tasks in Reason2Drive.}
    \label{fig-pie}
\end{minipage}
\hspace{4mm}
\begin{minipage}[b]{0.55\textwidth}
    \centering
    \resizebox{1.0\linewidth}{!}{
    \begin{tabular}{l|ccc|c}
    \midrule[1.0pt]
    \rowcolor[gray]{0.8}\diagbox[trim=l]{Target}{Task} & Perception (PE) & Prediction (PR) & Reasoning (RE) & Total \\
    \midrule[1.0pt]
    Ego vehicle & 22629 & 17173 & 18868 & 58670 \\
    Single object & 71882 & 61667 & 72006 & 205555 \\
    Multi objects & 102012 & 94502 & 52175 & 248689 \\
    Scenario & 49589 & 42795 & 27657 & 120041 \\
    \midrule[1.0pt]
    Total & 246112 & 216137 & 170706 & 632955 \\
    \midrule[1.0pt]
    \end{tabular}}
    \tabcaption{The statistics of different tasks in Reason2Drive dataset.}
    \label{tab-stat}
\end{minipage}
\vspace{-5mm}
\end{figure*}

\subsection{Dataset analysis} \label{dataset analysis}
Tab.~\ref{tab-bench} and Fig.~\ref{fig-scatter} demonstrate the comparison between our Reason2Drive dataset and existing benchmarks.
It is noteworthy that our benchmark stands as the largest dataset to date, surpassing others in terms of both dataset size and the inclusion of extensive long-text chain-based reasoning references.

To further investigate the property of Reason2Drive dataset, we count the distribution and sample numbers of our dataset in Fig.~\ref{fig-pie} and Tab.~\ref{tab-stat}.
We split the dataset according to the task and target.
The benchmark exhibits a balanced distribution, with multi-object tasks constituting the majority.
Single-object and scenario-level questions are of similar quantities.
The fewest questions are related to the ego-vehicle.
Additionally, perception, prediction and reasoning questions are distributed as 39\%, 34\%, and 27\%, respectively.
More dataset details are provided in the appendix.

\subsection{Benchmark protocol} \label{eval metric}
It is worth noting that previous works~\cite{qian2023nuscenes, malla2023drama, dewangan2023talk2bev} simply utilize metric scores widely used in the NLP community, including BLEU~\cite{papineni2002bleu}, CIDEr~\cite{vedantam2015cider} and METEOR~\cite{banerjee2005meteor}.
However, these metrics mainly measure text generation from a holistic perspective, without considering the causal relationship between the reasoning steps and the final conclusion.
While they perform well in translation and captioning, their efficacy is limited when it comes to reasoning.
Moreover, the existing evaluation system is confined to assessing the semantic quality of the results.
It fails to effectively evaluate the quality of the perceived results, a crucial aspect supporting automatic driving decision-making.
To address these dilemmas, inspired by \cite{yu2022alert} and \cite{golovneva2022roscoe}, we develop a novel evaluation protocol, ADRScore, to measure the correctness of the reasoning chains towards autonomous driving.

\noindent{\textbf{Preliminary.}}
To begin with, we denote the generated reasoning steps as hypothesis $\Vec{h}=\{h_1, ..., h_N\}$, and the gold annotation as reference $\Vec{r}=\{r_1, ..., r_K\}$.

At the core of reasoning metrics is the reasoning alignment vector from the $N$-step hypothesis $h$ to the $K$-step reference:

\begin{equation}
align(\Vec{h} \rightarrow \Vec{r}) = \{\alpha_1,...,\alpha_N\}, 
\end{equation}
where alignment value $\alpha_i$ represents the semantic similarity between the corresponding hypothesis step and the most similar reference step:
\begin{equation}
\begin{aligned}
\alpha_i = max_{j=1}^{K}s_{i,j}, \\
s_{i,j} = cos(h_i, r_j).
\end{aligned}
\end{equation}
$\alpha_i \in [0, 1]$ explicitly measures the grounding of the step-wise reasoning with respect to the reference, and $cos(\cdot)$ denotes the cosine similarity between the corresponding sentence embeddings, which are extracted a pre-trained bert model.
Based on the above reasoning alignment vector, we propose the following metrics to thoroughly measure the quality of reasoning steps.

\begin{figure*}[t]
    \begin{center}
        \includegraphics[width=1.0\linewidth]{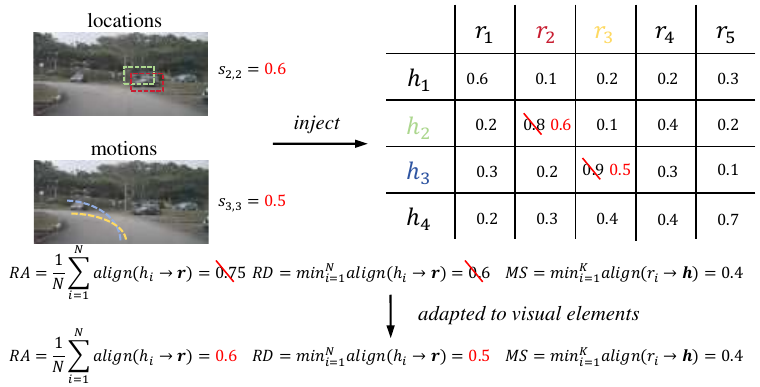}
    \end{center}
    \vspace{-4mm}
    \caption{Illustration of ADRScore and its visual adaptation. When substituting semantic similarities with actual geometric similarities, the score decreases.}
    \label{fig-metric}
    \vspace{-7mm}
\end{figure*}

\noindent{\textbf{Reasoning alignment.}}
The most straightforward way to evaluate the correctness of the hypothesis reasoning chain is to compare the degree of overlap between the hypothesis and the reference.
One way of doing that is to measure the reasoning alignment between them:
\begin{equation}
RA = \frac{1}{N} \sum_{i=1}^{N} align(h_i \rightarrow \Vec{r}).
\end{equation}

\noindent{\textbf{Redundancy.}}
To find chains that contain information that is not required to solve the problem (i.e., redundant steps), we identify those hypothesis steps that are least aligned with the reference steps.
This metric punishes the chain with steps that are not required for the correct solution.
\begin{equation}
RD = min_{i=1}^{N} align(h_i \rightarrow \Vec{r}).
\end{equation}

\noindent{\textbf{Missing step.}}
To identify steps that are missing from the hypothesis but could be required to solve the problem, we look at the alignment between reference and the hypothesis, similar to \textit{Redundancy}.
However, here we go through each step in the reference, and check if there is a similar step in the hypothesis:
\begin{equation}
MS = min_{i=1}^{K} align(r_i \rightarrow \Vec{h}).
\end{equation}
Finally, the aggregated metric score is the average of the above performance, which is:
\begin{equation}
ADRScore = \frac{1}{3} (RA + RD + MS).
\end{equation}

\noindent{\textbf{Adapted to visual elements.}}
To further adapt to the realistic driving process, we promote the above metric to the situation with visual elements, named ADRScore-S.
Specifically, as illustrated in Fig.~\ref{fig-metric}, when the hypothesis step $h_{i}$ and reference step $r_{j}$ contains visual elements, \ie, the locations and motions predicted for further reasoning, the similarity score becomes:

\begin{equation}
    s_{i,j} = \frac{\tau - M(h_{i}, r_{j})}{\beta},
\end{equation}
where $M(\cdot)$ measures the mean square error between two perceptual elements.
And $\tau$ and $\beta$ are used to normalize it to $[0, 1]$ to match the distribution of semantic-level similarity.
ADRScore-S more harshly measures the performance on spatial reasoning as the metric calculates the error of visual elements in spatial instead of text semantic.
The latter is too lenient for visual predictions in language.

\section{Methodology}
In this section we introduce our framework in Sec.~\ref{model arch}, followed by the training details provided in Sec.~\ref{training details}.

\subsection{Model architecture} \label{model arch}
We observe that most VLMs struggle to effectively handle object-level perceptual information, including the input of visual priors and predictions of object locations, which are indispensable in autonomous driving scenarios.
The limitation is primarily due to (i) the lack of a targeted tokenizer and (ii) decoder solely composed of a language model, resulting in subpar reasoning performance.

\begin{figure*}[t]
    \begin{center}
        \includegraphics[width=1.0\linewidth]{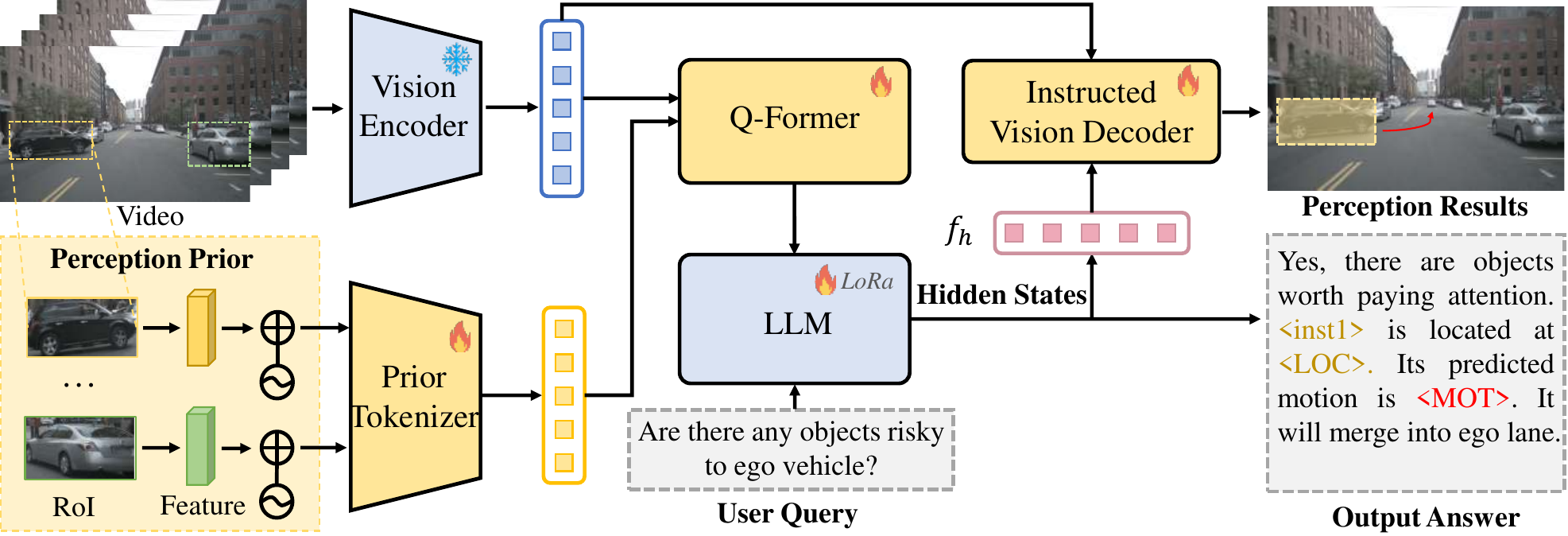}
    \end{center}
    \vspace{-4mm}
    \caption{The pipeline of our framework. The input video and perceptual priors are tokenized using the vision encoder and prior tokenizer. 
    A Q-former then aligns them to the text's feature space. 
    The LLM and instructed vision decoder predict answers with precise perception results for user queries.
    The highlighted yellow box and red curve in the perception results respectively represent the visualization of \texttt{<LOC>} and \texttt{<MOT>}.}
    \label{fig-framework}
    \vspace{-6mm}
\end{figure*}

To address this challenge, as illustrated in Fig.~\ref{fig-framework}, we introduce a straightforward yet effective framework that enhances existing VLMs with two new components: a prior tokenizer and an instructed vision decoder.
Notably, the design of the components is not for detection but to aid interpretable visual reasoning based on available perception inputs.
These modules aim to strengthen the capabilities of the model to utilize object-level perceptual elements in both extracting visual priors and generating perceptual predictions for visual reasoning.

\noindent\textbf{Vision encoder.}
Our model accepts both video frames and text inputs, along with perceptual priors, and tokenizes them into embeddings.
For a sequence of video frames $(V_1, V_2, ..., V_N)$, features are extracted using a pretrained Blip-2 visual encoder~\cite{li2023blip} $F_{v}$ and aggregated through concatenation:

\begin{equation}
    f_{v} = F_{v}(V_1) \oplus F_{v}(V_2) \oplus ... \oplus F_{v}(V_N).
\end{equation}

\noindent\textbf{Prior tokenizer.}
We propose a novel tokenization strategy tailored to taking advantage of visual cues.
The motivation is grounded in the acknowledgement that extracting and aligning visual features is considerably simpler and more suitable compared to compelling the LLM to comprehend ambiguous positional descriptions.
Direct textual input to the LLM may result in challenges such as information loss, as textual representation may not fully capture image details and context, especially in complex scenarios with dynamic object positions and velocities.
To tackle this issue, we design a novel tokenizer $F_{p}$, implemented as a two-layer MLP, to independently extract local image features and positional embeddings from visual priors:

\begin{equation}
    f_{p} = F_{p}(f_{r} + E(P)),
\end{equation}
where $f_{r}$ represents the region-level features extracted from the image-level features $f_{v}$ according to the precise locations of perception priors $P$.
These features are aligned to 7 $\times$ 7 size using the RoIAlign~\cite{dollar2017mask} operation and fused into a single embedding $f_{r}$.
And $E(\cdot)$ is a positional encoding function mapping the geometry locations and motions into the same dimension of $f_{r}$.

\noindent\textbf{LLM.}
After we tokenize the video and perception priors into embedding $f_{v}$ and $f_{p}$, a projector $Q$ (Q-former~\cite{li2023blip} in this work) is adopted to align the non-text features into textual domain:

\begin{equation}
    f_{q} = Q(f_{v}, f_{p}).
\end{equation}
Then, to generate the final text output, we utilize the LLM for further language processing with the extracted text embedding $f_{t}$:

\begin{equation}
    \hat{y}_{t} = F(f_{t}, f_{q}).
\end{equation}

\noindent\textbf{Instructed vision decoder.}
Current works~\cite{fu2023drive,deruyttere2022talk2car} treat the LLM as a versatile tool to generate answers and inferences without intermediate reasoning steps, let alone considering the perceptions of the agent toward driving scenes.
However, the perception ability of the agent towards driving scenarios is an indispensable part of a reliable driving procedure.
Moreover, recent works~\cite{lai2023lisa} have demonstrated that, rather than training with textualized perceptual sequences, incorporating the perception abilities into the multi-modal LLM brings a significant improvement.
To this end, inspired by \cite{lai2023lisa}, we integrate new perception capabilities into the multi-modal LLM.
Specifically, we expand the original LLM vocabulary by introducing new tokens as placeholders, denoted as \texttt{<LOC>} and \texttt{<MOT>}, to signify the request for the perception output.
When the LLM aims to generate a specific perception, the output $\hat{y}_{t}$ should include a designed token.
We then extract the last-layer textual features corresponding to the specific token and apply an MLP projection layer to obtain the hidden embedding $f_h$.
Finally, the textual embedding and visual features are fed into the instructed vision decoder to decode the predictions:

\begin{equation}
    \hat{P} = D(f_{v}, f_{h}).
\end{equation}
This module is comprised of a transformer decoder for features alignment~\cite{carion2020end} and task-specific heads designed to generate object locations and motions.

\subsection{Training details} \label{training details}
\noindent\textbf{Training objectives.}
The model is trained end-to-end using the text generation loss $\mathcal{L}_{txt}$ and the perception output loss $\mathcal{L}_{per}$:

\begin{equation}
    \mathcal{L} = \mathcal{L}_{txt} + \lambda_{per}\mathcal{L}_{per},
\end{equation}
where $\lambda_{per}$ is the balancing term.
Specifically, $\mathcal{L}_{txt}$ is the auto-regressive cross-entropy loss for text generation, and $\mathcal{L}_{per}$ encourages the instructed vision decoder to generate accurate locations and motions, which is similar to traditional detection loss and is employed with the combination of binary cross-entropy loss and MSE loss.
More details are included in the appendix.

\noindent\textbf{Tuning strategy.}
Our tuning strategy consists of two stages: the pre-training stage and the fine-tuning stage.
In the pre-training stage, we initialize the weights from instructBLIP~\cite{Dai2023InstructBLIPTG}, including the pre-trained vision encoder, Q-former and LLM, and freeze the parameters of LLM and vision tokenizer $F_{v}$.
We train the prior tokenizer $F_{p}$ and Q-former $Q$ to align visual priors with text, along with the instructed vision decoder $D$ to enhance visual localization capabilities.
The fine-tuning phase equips the LLM with reasoning abilities in autonomous driving using the instructed vision decoder.
To retain pre-trained LLM generalization, we employ efficient fine-tuning with LoRA~\cite{hu2021lora}.
The vision encoder and prior tokenizer $F_{p}$ remain fixed, while the instructed vision decoder $D$ is fully fine-tuned.
Word embeddings of the LLM and Q-former are also trainable.
\section{Experiments}
\begin{table*}[t]
\renewcommand\arraystretch{1.5}
\begin{center}
\caption{Results of different models on the Reason2Drive validation set. We evaluate with ADRScore as well as captioning-based metrics.}
\label{tab-reason}
\vspace{-3mm}
\resizebox{0.80\linewidth}{!}{
\begin{tabular}{c|c|cc|cccc}
\midrule[1.0pt]
\rowcolor[gray]{0.8}& & \multicolumn{2}{c|}{Reasoning metric} & \multicolumn{4}{c}{Captioning metric} \\
\rowcolor[gray]{0.8}\multirow{-2}{*}{Methods} & \multirow{-2}{*}{LLM} & ADRScore & ADRScore-S & B@4 & METEOR & ROUGE & CIDEr \\
\midrule[1.0pt]
\multirow{2}{*}{Blip-2~\cite{li2023blip}} & OPT-2.7B~\cite{zhang2022opt} & 0.296 & 0.162 & 0.361 & 0.249 & 0.443 & 0.174 \\
 & FlanT5-XL~\cite{chung2022scaling} & 0.310 & 0.171 & 0.368 & 0.256 & 0.451 & 0.187 \\
\midrule
\multirow{2}{*}{InstructBLIP~\cite{Dai2023InstructBLIPTG}}  & FlanT5-XL & 0.329 & 0.187 & 0.376 & 0.269 & 0.462 & 0.196 \\
& Vicuna-7B~\cite{peng2023instruction} & 0.351 & 0.214 & 0.408 & 0.294 & 0.484 & 0.211 \\
\midrule
MiniGPT-4~\cite{zhu2023minigpt} & Vicuna-7B & 0.338 & 0.203 & 0.396 & 0.286 & 0.475 & 0.219 \\
\midrule
\multirow{2}{*}{Ours} & FlanT5-XL & 0.457 & 0.420 & 0.451 & 0.349 & 0.520 & 0.292 \\
& Vicuna-7B & \textbf{0.463} & \textbf{0.432} & \textbf{0.457} & \textbf{0.356} & \textbf{0.529} & \textbf{0.298} \\
\midrule[1.0pt]
\end{tabular}}
\end{center}
\vspace{-7mm}
\end{table*}
\begin{table}[t]
\renewcommand\arraystretch{1.5}
\begin{center}
\caption{Ablations on different combinations of training tasks.}
\label{tab-task}
\vspace{-3mm}
\resizebox{0.8\linewidth}{!}{
\begin{tabular}{ccc|cc|cccc}
\midrule[1.0pt]
\rowcolor[gray]{0.8}\multicolumn{3}{c|}{Tasks} & \multicolumn{2}{c}{Reasoning metric} & \multicolumn{4}{c}{Captioning metric} \\
\rowcolor[gray]{0.8}Perception & Prediction & Reasoning & ADRScore & ADRScore-S & B@4 & METEOR & ROUGE & CIDEr \\
\midrule[1.0pt]
\cmark & & & 0.282 & 0.253 & 0.422 & 0.307 & 0.479 & 0.226 \\
\cmark & \cmark & & 0.297 & 0.264 & 0.419 & 0.310 & 0.479 & 0.228 \\
& & \cmark & 0.351 & 0.323 & 0.430 & 0.325 & 0.495 & 0.263 \\
\cmark & & \cmark & 0.407 & 0.364 & 0.435 & 0.337 & 0.501 & 0.274 \\
\rowcolor[gray]{0.9}\cmark & \cmark & \cmark & 0.463 & 0.432 & 0.457 & 0.356 & 0.529 & 0.298 \\
\midrule[1.0pt]
\end{tabular}}
\end{center}
\vspace{-8mm}
\end{table}
We benchmark various baseline models and present our method on Reason2Drive dataset.
Sec.~\ref{exp setting} covers implementation details.
We assess reasoning performance using our proposed metric in Sec.~\ref{results}, perform ablation studies in Sec.~\ref{ablation} and provide more ablations and qualitative results in the appendix.

\subsection{Experimental setting} \label{exp setting}
Our main experiments are carried out on the complete Reason2Drive benchmark.
The dataset is collected from three different source datasets: nuScenes~\cite{caesar2020nuscenes}, Waymo~\cite{sun2020scalability}, and ONCE~\cite{2021Once}.
It is divided into training and validation sets based on segments, with 70\% allocated to the training set and 30\% to the validation set, ensuring no overlap in scenes between them.
The validation set also contains all the declared tasks and has been augmented.
The input consists of 5 frames of cropped images with a size of 224$\times$224 pixels.
During training, we leverage the AdamW~\cite{loshchilov2017decoupled} optimizer with a weight decay of 0.01.
We adopt a cosine learning rate decay scheduler with a max value of 3e-4 and a linear warm-up for the first 1000 iterations.
The weight of perception loss $\lambda_{p}$ is set to 1.0.
The normalization parameters $\tau$ and $\beta$ are selected to be 15 and 10 after empirical practice.
Our models are trained with a batch size of 8 on 8 V100 GPUs.
The baseline models are fine-tuned following the official tuning strategy, with the inputs consistent with our approach.
Specifically, the perception priors are also provided as textual inputs to baseline models for fair comparison.

\begin{table}[t]
\renewcommand\arraystretch{1.5}
\setlength{\tabcolsep}{20pt}
\begin{center}
\vspace{-5mm}
\caption{The quality of predicted visual elements.}
\label{tab-visual}
\vspace{-3mm}
\resizebox{0.8\linewidth}{!}{
\begin{tabular}{c|c|ccc}
\midrule[1.0pt]
\rowcolor[gray]{0.8}Predictions & Metric & MiniGPT-4 & Kosmos-2~\cite{peng2023kosmos} & Ours \\
\midrule[1.0pt]
Bounding box & Accuracy $\uparrow$ & 0.723 & 0.745 & \textbf{0.806} \\
Trajectory & ADE $\downarrow$ & 2.334 & 2.563 & \textbf{1.875} \\
\midrule[1.0pt]
\end{tabular}}
\end{center}
\vspace{-10mm}
\end{table}
\begin{minipage}[t]{\textwidth}
 \begin{minipage}[t]{0.45\textwidth}
  \centering
  \makeatletter\def\@captype{table}\makeatother
  \caption{Ablations on visual input and perception priors.}
  \label{tab-receiver}
  \resizebox{1.0\textwidth}{!}{
   \begin{tabular}{cc|cc|cc}
   \midrule[1.0pt]
   \rowcolor[gray]{0.8}\multicolumn{2}{c|}{Visual features} & \multicolumn{2}{c|}{Perceptual priors} & & \\
   \rowcolor[gray]{0.8}image-level & video-level & region-level & positional & \multirow{-2}{*}{ADRScore} & \multirow{-2}{*}{ADRScore-S} \\
   \midrule[1.0pt]
   \cmark & & & & 0.414 & 0.379 \\
   & \cmark & & & 0.431 & 0.394\ \\
   & \cmark & \cmark & & 0.447 & 0.418 \\
   \rowcolor[gray]{0.9} & \cmark & \cmark & \cmark & 0.463 & 0.432 \\
   \midrule[1.0pt]
   \end{tabular}}
 \end{minipage}
 \begin{minipage}[t]{0.45\textwidth}
  \centering
  \makeatletter\def\@captype{table}\makeatother
  \caption{Ablations on different settings of instructed vision decoder.}
  \label{tab-operator}
  \resizebox{0.95\textwidth}{!}{
   \begin{tabular}{ccc|cc}
   \midrule[1.0pt]
   \rowcolor[gray]{0.8}Pre-train & text embedding & MLP & ADRScore & ADRScore-S \\
   \midrule[1.0pt]
   & & & 0.387 & 0.361 \\
   \cmark & & & 0.421 & 0.396 \\
   \cmark & \cmark & & 0.455 & 0.425 \\
   \rowcolor[gray]{.9}\cmark & \cmark & \cmark & 0.463 & 0.432 \\
   \midrule[1.0pt]
   \end{tabular}}
 \end{minipage}
\vspace{-5mm}
\end{minipage}

\subsection{Reasoning performance evaluation} \label{results}
As demonstrated in Tab.~\ref{tab-reason}, we evaluate both ADRScore and traditional caption-based performance of different models on our benchmark.
It is worth noting that our method outperforms others comprehensively in all metrics.
We also observe that, despite there is a correlation between ADRScore and traditional metrics, while on the other hand, the performance gap is more pronounced in our metrics, specially revealed by ADRScore-S.
The results further substantiate reasoning ambiguities in traditional metrics, constraining the differentiation in benchmarking model performance.
Specifically, models with varying reasoning capabilities exhibit minimal disparities when evaluated using traditional metrics.

\subsection{Ablation study} \label{ablation}
\noindent\textbf{Task contributions.}
To investigate the synergies between different tasks, we separate and evaluate various types of tasks independently.
As shown in Tab.~\ref{tab-task}, we train our model on different combinations of tasks.
Most notably, training on reasoning tasks plays the most important role, indicating the necessity of relevant reasoning data for instructional tuning.
Based on reasoning tasks, perception and prediction tasks additionally enhance the models for visual reasoning, showing specific improvements of 4.1\% and 6.8\%, respectively.

\noindent\textbf{Quality of predicted visual elements.}
To validate the quality of predicted perceptual elements, we also conduct experiments to evaluate the bounding boxes and trajectories separately, as shown in Tab.~\ref{tab-visual}.
To ensure a fair comparison, we implement Kosmos-2~\cite{peng2023kosmos} due to its advantageous grounding capabilities.
The model is fine-tuned on our dataset following the official strategy.
Experimental results confirm the high quality of our predicted visual elements, with particular emphasis on the accuracy of the trajectories.

\noindent\textbf{The effects of tokenizers.}
To verify the effectiveness of the tokenizers, we conduct ablation studies to pinpoint where the improvements come from in Tab.~\ref{tab-receiver}.
Visual features from single frame to multi-frame bring 1.5\% improvement in ADRScore-S.
Perceptual priors, \ie, region-level features and positional embeddings bring 2.4\% and 1.4\% development.

\noindent\textbf{The effects of instructed vision decoder.}
To verify the efficiency of our instructed vision decoder, we conduct an ablation study to compare it with other methods.
As demonstrated in Tab.~\ref{tab-operator}, pre-training and textual embedding bring the major contribution (3.5\% and 2.9\% in ADRScore-S).

\noindent\textbf{Downstream tasks.}
We are also interested in understanding how our benchmark will contribute to downstream tasks, such as predicting control signals.
Following the approach in ADriver-I~\cite{jia2023adriver}, we generate control signals on nuScenes and fine-tune InstructBLIP~\cite{Dai2023InstructBLIPTG} directly for planning signal prediction.
To ablate the influence of our dataset, we also pre-train models on Reason2Drive before fine-tuning with control signals.
The results presented in Tab.~\ref{tab-control} demonstrate the supportive effect of Reason2Drive on downstream planning tasks.

\begin{table}[t]
\renewcommand\arraystretch{1.5}
\setlength{\tabcolsep}{20pt}
\begin{center}
\caption{Evaluation of control signals. B: B@4. M: METHOR.}
\label{tab-control}
\vspace{-4mm}
\resizebox{0.8\linewidth}{0.1\textheight}{
\begin{tabular}{c|c|cccc}
\midrule[1.0pt]
\rowcolor[gray]{0.8}& & & & \multicolumn{2}{c}{RMSE} \\
\rowcolor[gray]{0.8}\multirow{-2}{*}{Method} & \multirow{-2}{*}{LLM} & \multirow{-2}{*}{B $\uparrow$} & \multirow{-2}{*}{M $\uparrow$} & Speed $\downarrow$ & Steer $\downarrow$ \\
\midrule[1.0pt]
\multicolumn{5}{l}{\textit{Directly fine-tuned with control signals:}} \\
\midrule[1.0pt]
InstructBLIP & Vicuna-7B & 0.166 & 0.201 & 3.743 & 5.926 \\
\midrule[1.0pt]
\multicolumn{5}{l}{\textit{Additionally pre-trained on Reason2Drive:}} \\
\midrule[1.0pt]
InstructBLIP & Vicuna-7B & 0.192 & 0.237 & 3.086 & 5.151 \\
Ours & Vicuna-7B & \textbf{0.213} & \textbf{0.269} & \textbf{2.842} & \textbf{4.866} \\
\midrule[1.0pt]
\end{tabular}}
\end{center}
\vspace{-11mm}
\end{table}
\section{Conclusion}
In summary, Large Vision-Language Models (VLMs) have sparked interest in autonomous driving for their advanced reasoning capabilities.
However, the absence of datasets explaining decision-making processes hinders progress.
To tackle this, we introduce Reason2Drive benchmark, comprising 600K+ video-text pairs for interpretable reasoning in complex driving scenarios.
It outperforms existing datasets in scale, sources and task diversities.
We also propose a novel evaluation protocol for chain-based reasoning, addressing existing semantic ambiguities.
To uncover insights into their reasoning abilities, our work evaluates various VLMs and proposes an efficient method to boost the ability of models to utilize object-level perceptual elements in both the encoder and decoder.
We expect our work could propel further advancements in interpretable reasoning for autonomous systems.

\section*{Acknowledgments}
This work was supported in part by National Natural Science Foundation of China (Grant No. 62106050 and 62376060),
Natural Science Foundation of Shanghai (Grant No. 22ZR1407500), 
USyd-Fudan BISA Flagship Research Program and Lingang Laboratory (Grant No. LG-QS-202202-07).

\bibliographystyle{splncs04}
\bibliography{main}

\appendix

\section{More Details of Reason2Drive}
\subsection{Words distribution}
\begin{figure*}[tbhp]
    \begin{center}
        \includegraphics[width=1.0\linewidth]{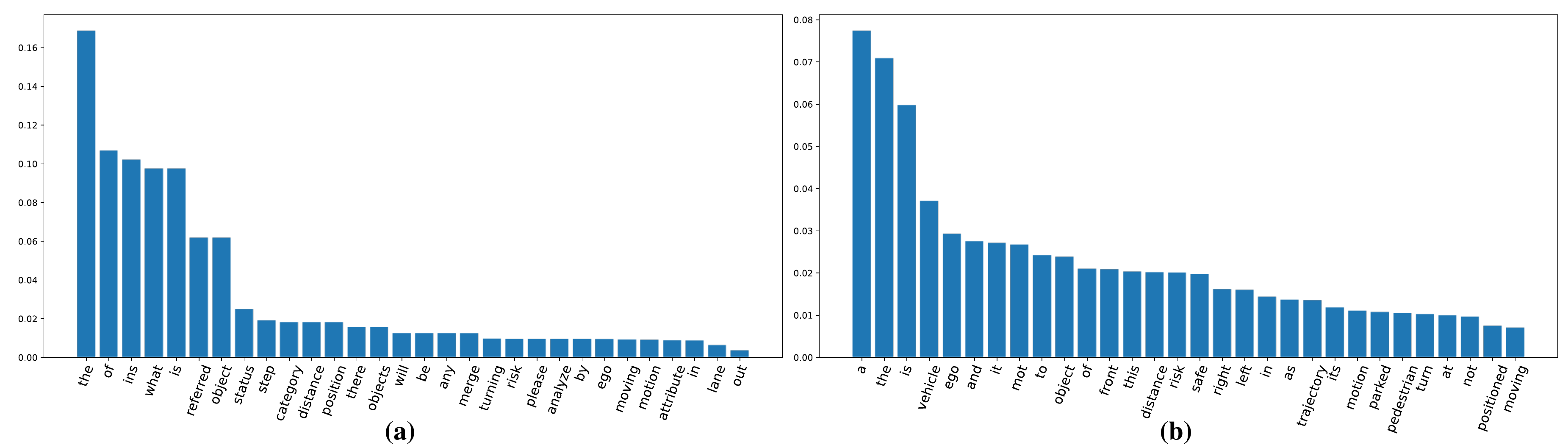}
    \end{center}
    \vspace{-4mm}
    \caption{Words distributions in (a) questions and (b) annotated answers.}
    \label{fig-word}
    \vspace{-7mm}
\end{figure*}

We count the distribution of the words, as is illustrated in Fig.~\ref{fig-word}.
From the words distribution, we can observe that Reason2Drive has a large range of words that describe perceptions, predictions and reasoning tasks, like ``moving'', ``distance'', and ``risk''.

\begin{table}[t]
\renewcommand\arraystretch{1.5}
\begin{center}
\resizebox{1.0\linewidth}{!}{
\begin{tabular}{c|c|c|c}
\midrule[1.0pt]
\rowcolor[gray]{0.8}Task & Sub-task & Target & Template \\
\midrule[1.0pt]
\multirow{15}{*}{Perception} & \multirow{3}{*}{Category} & Single object & What is the category of the referred object? \\
& & Multi objects & Are any of these objects vehicles? \\
& & Scenario & How many vehicles in the driving scenario? \\
\cmidrule{2-4}
& \multirow{3}{*}{Attribute} & Single object & What is the moving status of the referred object? \\
& & Multi objects & Which of these objects is stopped? \\
& & Scenario & Are there any objects parked in the driving scenario? \\
\cmidrule{2-4}
& \multirow{3}{*}{Distance} & Single object & What is the distance of the referred object towards ego? \\
& & Multi objects & Which of these objects is closest to the ego? \\
& & Scenario & Are there any objects too close to the ego in the driving scenario? \\
\cmidrule{2-4}
& \multirow{3}{*}{Position} & Single object &  What is the position of the referred object? \\
& & Multi objects & Which of these objects is located at left of the ego? \\
& & Scenario & Are there any objects right in front of the ego in the driving scenario? \\
\midrule
\multirow{14}{*}{Prediction} & \multirow{2}{*}{Motion} & Single object & What is the future trajectory of the referred object? \\
& & Ego & What is the future trajectory of the ego vehicle? \\
\cmidrule{2-4}
& \multirow{3}{*}{Moving strategy} & Single object & What will the moving status of the referred object be in a few seconds? \\
& & Multi objects & Which of these objects will be stopped in a few seconds? \\
& & Ego & What will the moving status of the ego vehicle be in a few seconds? \\
\cmidrule{2-4}
& \multirow{3}{*}{Turn} & Single object & Which direction will the referred object turn? \\
& & Multi objects & Which of these objects will turn left? \\
& & Scenario & Will there be any objects turning right in the driving scenario? \\
\cmidrule{2-4}
& \multirow{3}{*}{Trend} & Single object & Will the referred object approach or stay away? \\
& & Multi objects & Which of these objects will approach? \\
& & Scenario & Will there be any objects approaching the ego vehicle? \\
\cmidrule{2-4}
& \multirow{3}{*}{Merge} & Single object & Will the referred object merge in/out of the ego lane? \\
& & Multi objects & Which of these objects will merge in/out of the ego lane? \\
& & Scenario & Will there be any objects merging in/out of the ego lane? \\
\midrule
\multirow{6}{*}{Reasoning} & \multirow{2}{*}{Driving strategy} & Single object & What is the referred object doing and what causes it? \\
& & Ego & What is the ego vehicle doing and what causes it? \\
\cmidrule{2-4}
& \multirow{2}{*}{Risk} & Single object & Is the referred object risky to the ego vehicle's normal driving? \\
& & Scenario & Is there any risk to the ego vehicle's normal driving in the scenario? \\
\cmidrule{2-4}
& \multirow{2}{*}{Control} & Single object & What will the referred object do in a few seconds for safety driving and why? \\
& & Ego & What will the ego vehicle do in a few seconds for safety driving and why? \\
\midrule[1.0pt]
\end{tabular}}
\end{center}
\caption{Details of sub-tasks and question templates.}
\label{tab-subtask}
\end{table}

\subsection{Detailed sub-tasks in Reason2Drive}
In this section, we present more dataset details about Reason2Drive.
As introduced in the main paper, we divide the autonomous driving tasks to three distinct groups to acquire diversified data: perception, prediction and reasoning.
In detail, we have a further breakdown of driving tasks, covering 15 sub-tasks for perception, 14 for prediction and 6 for reasoning, with specific examples provided in Tab.~\ref{tab-subtask}.

\subsection{Prompts and human instructions}
\begin{figure*}[t]
    \begin{center}
        \includegraphics[width=1.0\linewidth]{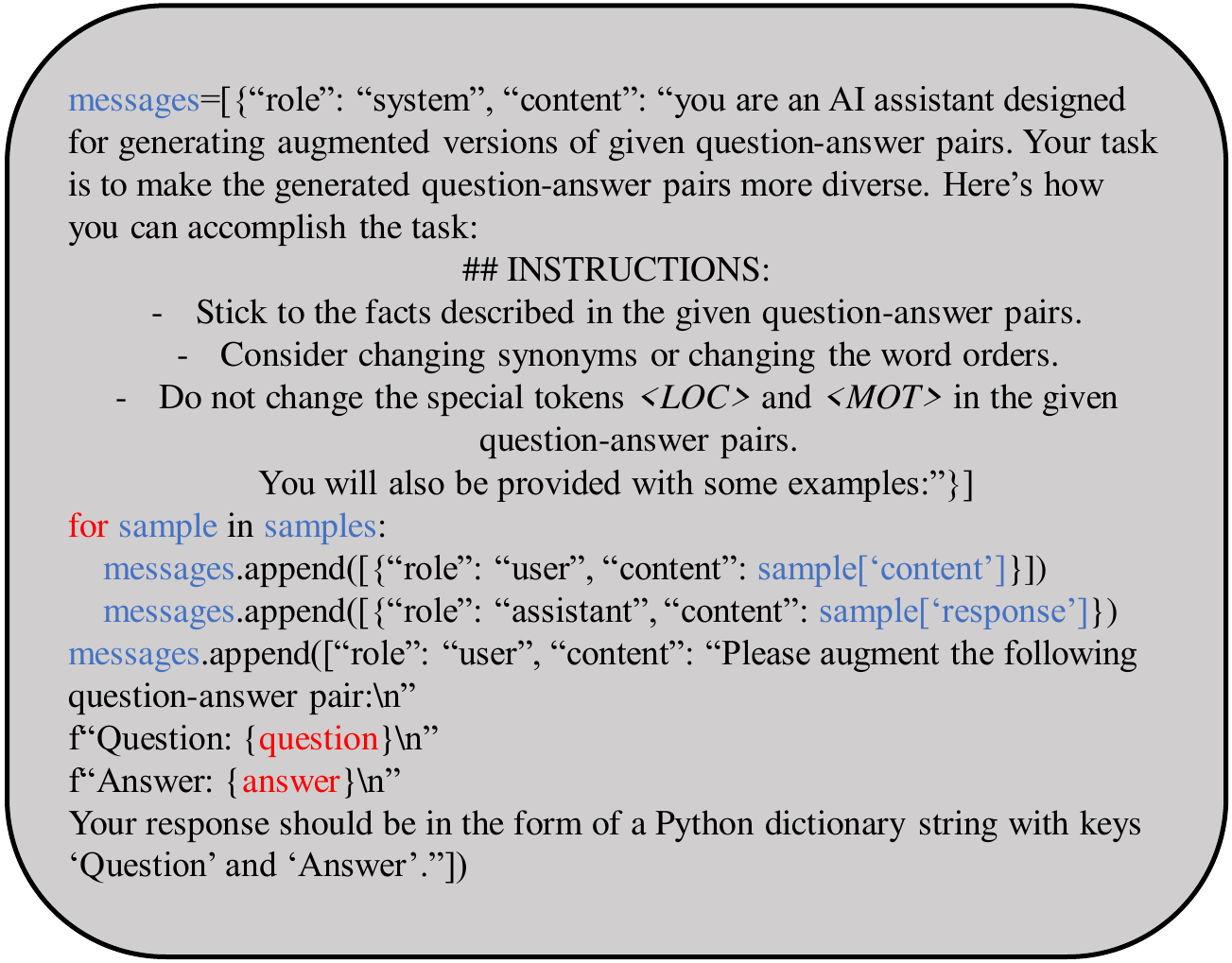}
    \end{center}
    \vspace{-4mm}
    \caption{Prompts and human instructions on augmenting question-answer pairs.}
    \label{fig-gpt}
    \vspace{-3mm}
\end{figure*}

In Fig~\ref{fig-gpt}, we show the prompts and human instructions for generating augmented question-answer pairs.
We provide system prompts for GPT of being an AI assistant designed for augmenting question-answer pairs.
For each sample in the given examples, the ``content" has the exemplar question-answer pairs, and the ``response" refers to human-written instructions for demonstration.
Finally, the real question-answer pairs are provided in the user's content.

\section{More Implementation Details}
\subsection{Architecture}
For the frozen visual encoder, we employ ViT-G/14 from EVA-CLIP~\cite{sun2023eva} in the main paper, which is a state-of-the-art pre-trained vision transformer models.
We remove the last layer of the ViT and uses the second last layers' output features.

For the language model, we explore two types of LLMs: encoder-decoder-based LLMs and decoder-based LLMs.
For encoder-decoder-based LLMs, we employ FlanT5-XL~\cite{chung2022scaling}, which is an instruction-tuned model based on the encoder-decoder Transformer T5~\cite{xue2020mt5}.
For decoder-based LLMs, we select Vicuna~\cite{chiang2023vicuna}, a recently released decoder-only Transformer instruction-tuned from LLaMA~\cite{touvron2023llama}.

\subsection{Training loss}
Our model is trained with a language modelling loss $\mathcal{L}_{txt}$, where the task of the frozen LLM is to generate text conditioned on the extracted modality features of the Q-former.
Furthermore, we employ an auxiliary perception loss $\mathcal{L}_{per}$ to enhance the perceptual capability.
Specifically, a linear combination of a binary cross-entropy loss for classification and a regression loss is defined:
\begin{equation}
    \mathcal{L}_{per}(P,\hat{P}) = -\sum_{i=1}^{N}log\hat{P}_{c,i} + \lambda_{reg}\sum_{i=1}^{N}\mathcal{L}_{reg}(P_{b,i}, \hat{P}_{b,i}),
\end{equation}
where $\hat{P}_{c,i}$ and $\hat{P}_{b,i}$ are predicted classification and regression results of $\hat{P}$.
Loss function $\mathcal{L}_{reg}$ is employed by a MSE loss.
In practice, we select $\lambda_{reg}$ to be 0.25 as the balance term as a common setting in object detection tasks.

\subsection{Implementation of baseline models}
The baseline models are implemented following the official implementation and fine-tuned on Reason2Drive, with inputs consistent with our approach.
Specifically, perception priors are provided as textual inputs to the baseline models to ensure a fair comparison.
Since baseline models lack a vision decoder, we prompt the fine-tuned baseline models to output perceptual results in textual form.
These results can be identified using regular expressions for the evaluation of ADRScore-S.

\begin{table}[t]
\renewcommand\arraystretch{1.5}
\begin{center}
\resizebox{0.8\linewidth}{!}{
\begin{tabular}{c|c|cc}
\midrule[1.0pt]
\rowcolor[gray]{0.8}Method & Visual encoder & ADRScore & ADRScore-S \\
\midrule[1.0pt]
\multirow{2}{*}{Blip-2} & ViT-L/14~\cite{radford2021learning} & 0.294 & 0.155 \\
& ViT-G/14~\cite{sun2023eva} & 0.310 & 0.171 \\
\midrule
\multirow{2}{*}{InstructBLIP} & ViT-L/14 & 0.327 & 0.187 \\
& ViT-G/14 & 0.351 & 0.214 \\
\midrule
\multirow{2}{*}{Ours} & ViT-L/14 & 0.435 & 0.397 \\
& ViT-G/14 & \textbf{0.463} & \textbf{0.432} \\
\midrule[1.0pt]
\end{tabular}}
\end{center}
\vspace{-2mm}
\caption{Ablations on visual encoders.}
\label{tab-encoder}
\vspace{-7mm}
\end{table}

\section{More Ablations}
\subsection{Ablation of visual encoders}
We ablate the effects of employed visual encoders in Tab.~\ref{tab-encoder}.
For comparison, we explore two types of visual encoders: ViT-L/14 in CLIP~\cite{radford2021learning} and ViT-G/14 in EVA-CLIP~\cite{sun2023eva}.
We can draw the conclusion that the performance of visual encoder inevitably influences the VLMs especially in strict reason metric.

\subsection{Evaluated by GPT-4}
To validate the rationality of our reasoning scores, following~\cite{fu2023drive}, we employ GPT-4 to validate the generated answers in Tab.~\ref{tab-gpt}.
We can draw the conclusion that our method still achieves superior performance, which also indicates the rationality of our proposed metric.
In addition, our metrics are not gpt-dependent, thus they are not affected by the migration of GPT.
As a contrast, in addition to being more interpretable, our metrics are not affected by the migration of GPT, ensuring the independence and stability.
\begin{table}[t]
\renewcommand\arraystretch{1.5}
\begin{center}
\resizebox{0.8\linewidth}{0.1\textheight}{
\begin{tabular}{c|c|cc|cc}
\midrule[1.0pt]
\rowcolor[gray]{0.8}Methods & LLM & ADRScore & ADRScore-S & GPT-3.5 & GPT-4 \\
\midrule[1.0pt]
Blip-2 & OPT-2.7B & 0.450 & 0.332 & 0.479 & 0.458 \\
InstructBLIP & FlanT5-XL & 0.489 & 0.377 & 0.532 & 0.501 \\
MiniGPT-4 & Vicuna-7B & 0.469 & 0.352 & 0.519 & 0.467 \\
\midrule
Ours & Vicuna-7B & \textbf{0.593} & \textbf{0.561} & \textbf{0.643} & \textbf{0.628} \\
\midrule[1.0pt]
\end{tabular}}
\end{center}
\vspace{-2mm}
\caption{Evaluation results given by prompted ChatGPT.}
\label{tab-gpt}
\vspace{-7mm}
\end{table}

\subsection{Generalization}
To validate the method's generalization, we trained on the Reason2Drive benchmark with only the nuScenes dataset and tested on Waymo and ONCE in Tab.~\ref{tab-gen}.
We split the Reason2Drive benchmark into two sets, nuScenes (noted as N) and Waymo + ONCE (noted as W + O).
Compared with others, our method suffers limited performance drops.
The generalization results suggest that the world knowledge of LLM helps the model generalized to the unseen scenarios.
Thereby we observe that there is no significant gap between training from different sources.
\begin{table}[t]
\renewcommand\arraystretch{1.5}
\begin{center}
\resizebox{0.6\linewidth}{!}{
\begin{tabular}{c|c|c|cc}
\midrule[1.0pt]
\rowcolor[gray]{0.8}& & & \multicolumn{2}{c}{Testing} \\
\rowcolor[gray]{0.8}\multirow{-2}{*}{Method}& \multirow{-2}{*}{LLM} & \multirow{-2}{*}{Training} & N & W + O \\
\midrule[1.0pt]
\multirow{2}{*}{Blip-2} & \multirow{2}{*}{OPT-2.7B} & N & 0.205 & 0.104 \\
 &  & W + O & 0.183 & 0.121 \\
\midrule
\multirow{2}{*}{InstructBLIP} & \multirow{2}{*}{FlanT5-XL} & N & 0.255 & 0.116 \\
 &  & W + O & 0.212 & 0.155 \\
\midrule
\multirow{2}{*}{MiniGPT-4} & \multirow{2}{*}{Vicuna-7B} & N & 0.263 & 0.130 \\
 &  & W + O & 0.226 & 0.172 \\
\midrule
\multirow{2}{*}{Ours} & \multirow{2}{*}{Vicuna-7B} & N & \textbf{0.443} & \textbf{0.385} \\
 &  & W + O & \textbf{0.428} & \textbf{0.397} \\
\midrule[1.0pt]
\end{tabular}}
\end{center}
\vspace{-2mm}
\caption{Generalization ability when transferred to different sources of datasets. ADRScore-S is reported.}
\label{tab-gen}
\vspace{-5mm}
\end{table}

\section{Qualitative Examples}
\begin{figure*}[t]
    \begin{center}
        \includegraphics[width=1.0\linewidth]{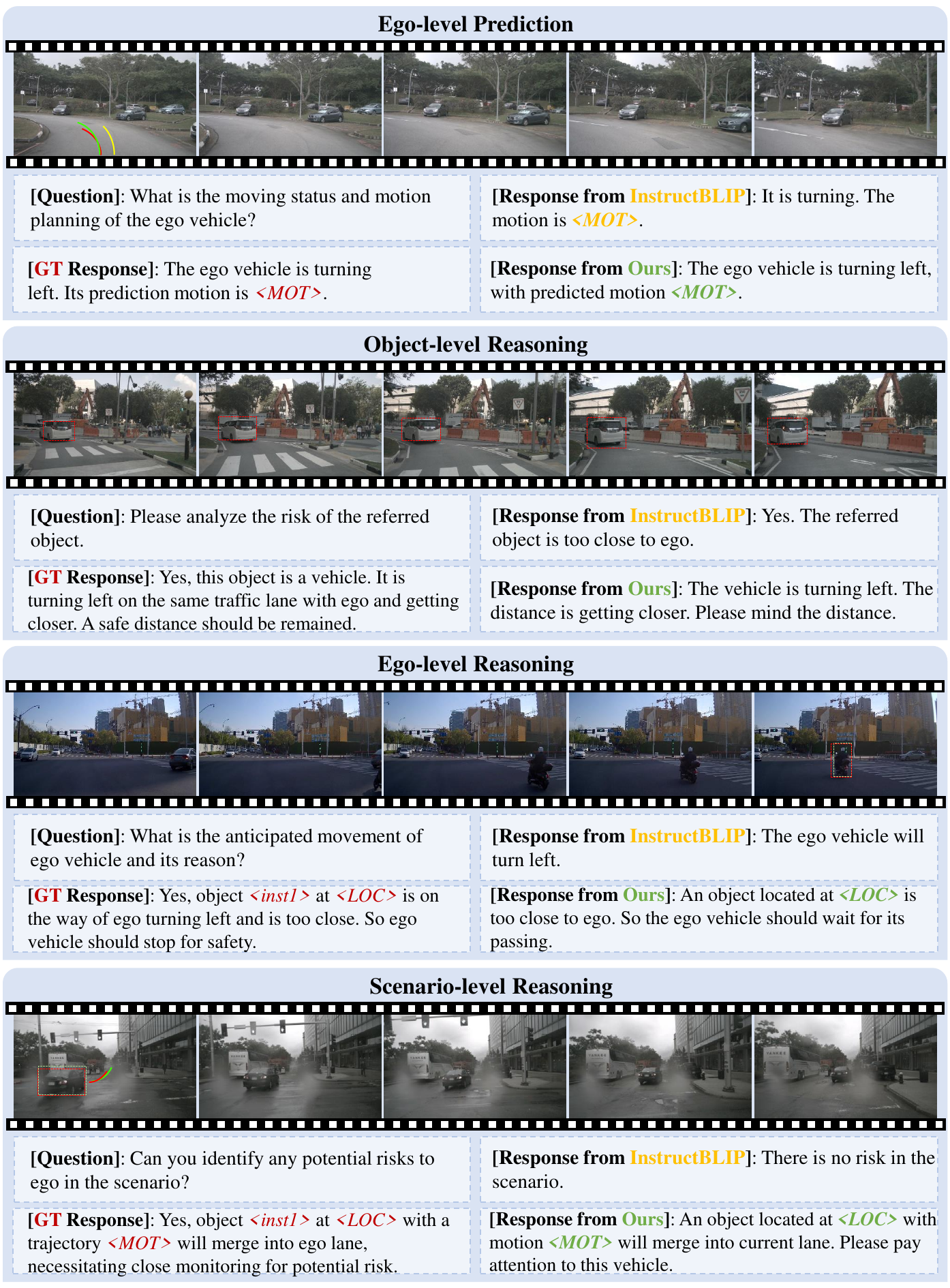}
    \end{center}
    \vspace{-4mm}
    \caption{Successful examples. Locations and motions are pictured in the first frame for better visualization. Ground truth in red color and prediction in green color.}
    \label{fig-success}
    \vspace{-3mm}
\end{figure*}
\begin{figure*}[t]
    \begin{center}
        \includegraphics[width=1.0\linewidth]{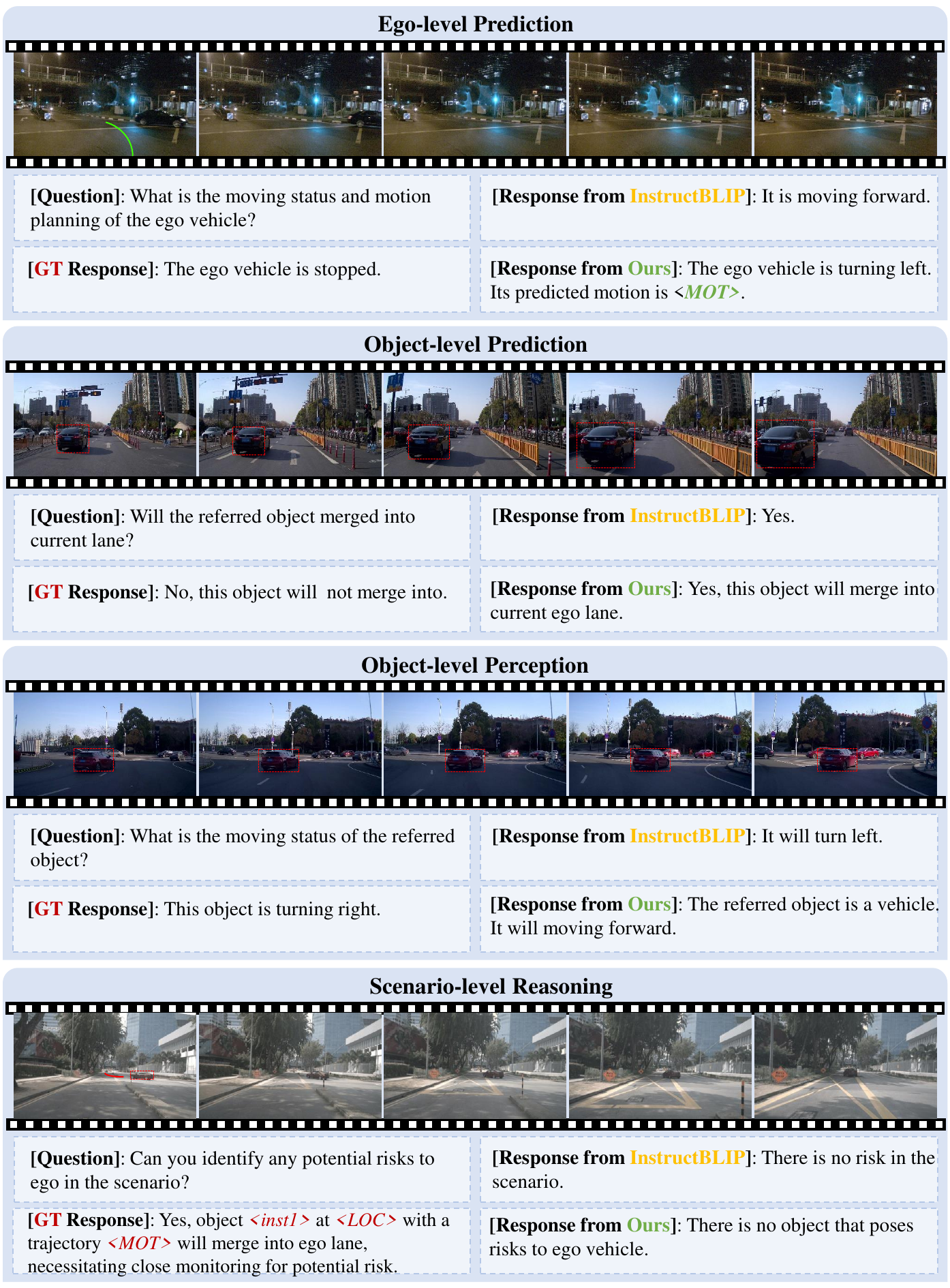}
    \end{center}
    \vspace{-4mm}
    \caption{Failure examples. Locations and motions are pictured in the first frame for better visualization. Ground truth in red color and prediction in green color.}
    \label{fig-fail}
    \vspace{-3mm}
\end{figure*}
\subsection{Successful Cases}
In Fig.~\ref{fig-success}, we visualize some of the successful cases in our Reason2Drive validation set.
In general, our method behaves better than InstructBLIP~\cite{Dai2023InstructBLIPTG} in most scenarios.
Our method performs well on the planning prediction of objects, the recognition of potential risks and reasoning steps under different levels of tasks.
The qualitative results demonstrate the effectiveness of our method towards interpretable and chain-based reasoning, which has great implications for autonomous driving.

\subsection{Failure Cases}
In Fig.~\ref{fig-fail}, we show the generation failures.
For some relatively complex driving scenarios, the existing methods, including ours, still make some mistakes.
In the first case of ego-level prediction, the network predicted the stooped ego vehicle to be turning because the slightly movement of the ego vehicle.
In the second and third cases of object-level perception and prediction, both our method and InstructBLIP misjudged the moving status of the referred object due to the relative displacement of the ego car.
Besides, the VLMs seem likely to miss recognition when opposed to distant risk objects, as illustrated in the fourth case.
These issues may be mitigated by targeted research to enhance the features of distance objects and the encoding of dynamic displacement of the ego vehicle in the future.

\end{document}